\documentclass[conference]{IEEEtran}
\IEEEoverridecommandlockouts

\usepackage{url}
\usepackage[colorlinks,linkcolor=red, citecolor=cyan]{hyperref}  
\usepackage{cite}
\usepackage{amsmath,amssymb,amsfonts}
\usepackage{algorithmic}
\usepackage{textcomp}
\usepackage{xcolor}
\usepackage{tikz}
\usepackage{multirow}
\usepackage{booktabs}
\usepackage{marvosym}
\usepackage{ifsym}
\usepackage{graphicx}
\usepackage{flushend}
\usepackage{subcaption}

\def\x{{\mathbf x}}

\def\z{{\mathbf z}}

\def\vx{{\vec{\mathbf x}}}

\def\vz{{\vec{\mathbf z}}}

\def\BibTeX{{\rm B\kern-.05em{\sc i\kern-.025em b}\kern-.08em
    T\kern-.1667em\lower.7ex\hbox{E}\kern-.125emX}}

\begin{document}

\title{Regularization-Based Efficient Continual Learning in Deep State-Space Models
\thanks{
This work was supported by the Shenzhen Science and Technology Program under Grant No. JCYJ20220530143806016, and in part by NSFC under Grant No. 62271433.  Feng Yin (\textit{yinfeng@cuhk.edu.cn}) is the Corresponding Author.}
}

\author{
	\IEEEauthorblockN{Yuanhang Zhang\textsuperscript{$ \dagger$},
	Zhidi Lin\textsuperscript{$\dagger$},
	Yiyong Sun\textsuperscript{$\dagger$},
	Feng Yin\textsuperscript{$\dagger$}(\textrm{\Letter}),
        and
        Carsten Fritsche\textsuperscript{$\ddagger$}
        }
	\IEEEauthorblockA{
	        $\dagger$  School of Science and Engineering, The Chinese University of Hong Kong, Shenzhen, China \\
                $\ddagger$  IAV GmbH, Weimarer Str. 10, 80807 Muenchen, Germany
	    }
	}

\maketitle

\begin{abstract}
Deep state-space models (DSSMs) have gained popularity in recent years due to their potent modeling capacity for dynamic systems. However, existing DSSM works are limited to single-task modeling, which requires retraining with historical task data upon revisiting a forepassed task. To address this limitation, we propose continual learning DSSMs (CLDSSMs), which are capable of adapting to evolving tasks without catastrophic forgetting. Our proposed CLDSSMs integrate mainstream regularization-based continual learning (CL) methods, ensuring efficient updates with constant computational and memory costs for modeling multiple dynamic systems. We also conduct a comprehensive cost analysis of each CL method applied to the respective CLDSSMs, and demonstrate the efficacy of CLDSSMs through experiments on real-world datasets. The results corroborate that while various competing CL methods exhibit different merits, the proposed CLDSSMs consistently outperform traditional DSSMs in terms of effectively addressing catastrophic forgetting, enabling swift and accurate parameter transfer to new tasks. 
\end{abstract}


\section{Introduction} \label{sec: introduction}
State-space models (SSMs) provide a fundamental framework for system identification and state inference in dynamic systems \cite{sarkka2013bayesian}. With their excellent model interpretability, diverse SSMs have found successful applications across various domains in recent decades, including robotic control, healthcare, climate change tracking and indoor positioning \cite{zhao2019cramer,xie2020learning,zhao2015particle,zhao2018sequential,lin2022output,lin2023ensemble,lin2023towards, lin2024towardse,alaa2019attentive}. Nevertheless, classic SSMs generally require prior knowledge of the underlying system dynamics, which are challenging to determine in advance. Consequently, there is a need to learn the dynamics from observed noisy measurements, giving rise to the development of data-driven SSMs \cite{lin2022output,lin2023ensemble,lin2023towards, lin2024towardse,krishnan2017structured, alaa2019attentive}.

One of the popular data-driven SSMs is deep state-space models (DSSMs), which leverage deep neural networks as their central modeling component, augmenting the learning and inference capabilities of classic SSMs and reducing the reliance on the modeling prior knowledge \cite{chung2015recurrent,krishnan2017structured,klushyn2021latent,gedon2021deep,karl2016deep}. Due to these advantageous properties, significant strides have been achieved in DSSMs, encompassing applications such as time series prediction\cite{chung2015recurrent,krishnan2017structured,klushyn2021latent}, nonlinear system identification\cite{masti2021learning, gedon2021deep,karl2016deep}, and healthcare \cite{krishnan2017structured, alaa2019attentive}. 

However, despite the notable performance of DSSMs in isolated tasks, their practical usage in real-world scenarios is likely to be hindered by limitations in memory and computing resources. This is because practical applications often involve successive arriving data, which further exacerbates the resource constraints \cite{yin2017distributed}. These constraints necessitate DSSMs to exhibit continual learning (CL) capabilities. For instance, in domains like autonomous driving systems\cite{yin2020fedloc}, where DSSMs are expected to navigate in dynamic environments, and continuously track without frequent model retraining or extensive storage of historical task data. Consequently, the development of DSSMs capable of continual learning becomes a critical step to quickly adapt to and learn from multiple tasks \cite{de2021continual}.
It is noted that, unlike multi-task learning, which jointly addresses multiple offline tasks \cite{zhang2021survey}, and transfer learning, which involves transferring knowledge from one task to another \cite{zhuang2020comprehensive}, continual learning emphasizes a model's capacity to learn continuously from sequential data streams. For further insights, one can refer to, e.g., \cite{chen2018lifelong}.

Recent research on continual learning predominantly concentrates on supervised learning and can be broadly classified into three main approaches: replay \cite{rolnick2019experience, rebuffi2017icarl,isele2018selective}, regularization  \cite{kirkpatrick2017overcoming,schwarz2018progress,aljundi2018memory,zenke2017continual,li2017learning,titsias2019functional,nguyen2017variational}, and parameter isolation \cite{xu2018reinforced}. More specifically, replay methods involve replaying a subset of previous data, regularization methods employ regularization terms to consolidate historical knowledge, and parameter isolation methods enable the learning model to develop new branches for future tasks while preserving parameters for previous tasks. 

With our aim to reduce the storage and computational overhead of DSSMs in continual learning, in this paper, we focus on the regularization-based approaches to eliminate the need for storing raw inputs akin to the replay methods. In addition, the regularization-based approaches can preserve the conciseness of the deep neural network in DSSMs, rendering a memory-efficient learning framework compared to the parameter isolation approaches. Our main contributions are summarized as follows:  
\begin{itemize}
    \item 
     We incorporate regularization-based continual learning methods into DSSMs, resulting in continual learning DSSMs (CLDSSMs). The proposed CLDSSMs demonstrate the capability to continually learn multiple dynamic systems without encountering catastrophic forgetting issues, rendering them adaptable to a wider range of system modeling applications. To the best of our knowledge, this paper marks the first exploration of continual learning in the context of DSSMs. 
    \item 
    We demonstrate that the CLDSSM, enhanced with various prevalent regularization-based continual learning methods, maintains a constant memory cost even with a continually expanding volume of data. Our approach also achieves superior results without requiring training with historical data, thus addressing the limitation of the standard DSSM. Furthermore, the continual learning methods employed in CLDSSMs acquire knowledge from historical tasks and thus can help expedite the training when encountering new related tasks, enabling the model to attain satisfactory results in earlier training phases.
    \item 
    We evaluate the performance of the proposed CLDSSMs on real-world datasets, showcasing their efficacy in overcoming catastrophic forgetting while achieving savings in both computational and memory costs, and enhancing model training. These results highlight the versatility of the CLDSSMs and their applicability to various real-world dynamic system modeling applications.
\end{itemize}
The subsequent sections of this paper are structured as follows. Some preliminaries and background about DSSMs are presented in Section \ref{sec: preliminary}. Section \ref{sec: proposed_mothod} outlines the pipeline of the proposed CLDSSMs. Numerical results are presented in Section \ref{sec: experiment}, and Section \ref{sec: conclusion} concludes this paper.

\section{Preliminaries}  \label{sec: preliminary}
This section begins with a brief introduction to DSSMs in Section \ref{sec: DSSMs}. Subsequently, Section \ref{subsec:enkf} reviews a learning method based on the autodifferentiable ensemble Kalman filter (EnKF) \cite{chen2022autodifferentiable}.

\subsection{Deep State-Space Models} \label{sec: DSSMs}

\begin{figure}[t!]
	\centering
	\footnotesize
	\begin{tikzpicture}[align = center, latent/.style={circle, draw, text width = 0.4cm}, observed/.style={circle, draw, fill=gray!20, text width = 0.4cm}, transparent/.style={circle, text width = 0.4cm}, node distance=1.1cm]
		\node[latent](z0) {${\z}_0$};
		\node[latent, right of=z0](z1) {${\z}_{1}$};
		\node[latent, right of=z1](z2) {${\z}_{2}$};
        \node[transparent, right of=z2](z3) {$\cdots$};
		\node[latent, right of=z3](zt-1) {$\!\!{\z}_{t-1}\!\!$};
		\node[latent, right of=zt-1](zt) {${\z}_{t}$};
		\node[transparent, right of=zt](zinf) {$\cdots$};
		\node[observed, below of=z1](x1) {${\x}_{1}$};
		\node[observed, below of=z2](x2) {${\x}_{2}$};
  	\node[transparent, right of=x2](x3) {$\cdots$};
		\node[observed, below of=zt-1](xt-1) {$\!\!{\x}_{t-1}\!\!$};
		\node[observed, right of=xt-1](xt) {${\x}_{t}$};
		\node[transparent, right of=xt](xinf) {$\cdots$};
        \draw[-latex] (z0) -- (z1);
		\draw[-latex] (z1) -- (z2);
  	\draw[-latex] (z1) -- (x1);
        \draw[-latex] (z2) -- (x2);
		\draw[-latex] (z2) -- (z3);
  	\draw[-latex] (z3) -- (zt-1);
        \draw[-latex] (zt-1) -- (xt-1);
		\draw[-latex] (zt-1) -- (zt);
		\draw[-latex] (zt) -- (xt);
        \draw[-latex] (zt) -- (zinf);

	\end{tikzpicture}
	\caption{Graphical representation of an SSM. }
	\label{fig:graphical_model}
\end{figure}
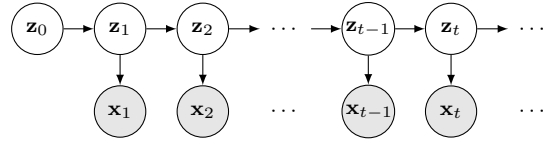

As depicted in Fig.~\ref{fig:graphical_model}, we consider an SSM that characterizes the probabilistic relationship between the latent state $\z_t \!\in\! \mathbb{R}^{d_z}$ and the observation $\x_t \!\in\! \mathbb{R}^{d_x}$, as expressed by the following equations:
\begin{align}
    &\z_{t} = F_\alpha(\z_{t-1}) + \xi_{t},  \qquad \xi_{t} \sim \mathcal{N}(0, Q_{\beta}), \label{eq:ssm1} \\
    &\x_{t} = H\z_{t} + \eta_{t},  \qquad \qquad \eta_{t} \sim \mathcal{N}(0, R), \label{eq:ssm2}
\end{align}
where $F_\alpha(\cdot)$ is the transition function that maps the latent state $\z_{t-1}$ to the future state $\z_t$, with $1\leq t \leq T$. The emission function (see Eq.~\eqref{eq:ssm2}) that maps latent states to observations is assumed to be linear and known, with the coefficient matrix $H\in\mathbb{R}^{{d_x}\times{d_z}}$. The noises $\xi_{t}$ and $\eta_{t}$ are additive and independent Gaussian random variables. The model parameters, denoted by $\theta \triangleq \{\alpha, \beta\}$, are both unknown and time-invariant. 
When the transition function is modeled using deep neural networks, the resulting model is referred to as a deep SSM (DSSM) \cite{gedon2021deep}. Within DSSM, $\alpha$ represents the parameter associated with the transition neural network, while $\beta$ is the covariance matrix of the Gaussian noise $\xi_{t}$.

In DSSMs, one of the most challenging tasks is to simultaneously learn the model parameter $\theta$ and infer the latent state of interest $\z_t$. This typically involves dealing with the model marginal likelihood \cite{theodoridis2020machine,cheng2022rethinking}, which can be expressed mathematically as:
\begin{equation}
    p_{\theta}(\vx) = \int p_{\theta}\left(\vx, \vz \right) \mathrm{d} \vz = \int p(\vx |\vz)p_{\theta}(\vz) \mathrm{d} \vz,
    \label{eq:model_evidence}
\end{equation}
where $\vx \triangleq \{\x_t\}_{t=1}^T$ and $\vz \triangleq \{\z_t\}_{t=1}^T$ represent the sequences of observations and latent states of length $T$, respectively.  
The term $p_{\theta}(\vz)$ is the prior distribution of the latent states, and $p(\vx|\vz)$ represents the model emission or likelihood function.
Since the integral in Eq.~\eqref{eq:model_evidence} is generally intractable, further approximation is required.

\subsection{Autodifferentiable Ensemble Kalman Filters} \label{subsec:enkf}
Numerous approximation techniques exist for addressing the intractable model evidence in Eq.~\eqref{eq:model_evidence}, including variational inference, MCMC sampling, Laplace, among others \cite{theodoridis2020machine, cheng2022rethinking, chung2015recurrent,krishnan2017structured,klushyn2021latent, masti2021learning, gedon2021deep,karl2016deep}. This paper predominantly employs a very recent approximation approach based on the ensemble Kalman filter (EnKF) \cite{evensen2003ensemble,roth2017ensemble}, namely the autodifferentiable EnKF \cite{chen2022autodifferentiable}. The autodifferentiable EnKF leverages the well-established EnKF for state inference while exploiting the autodifferentiation feature to simultaneously learn the parameter $\theta$. Specifically, to derive the marginal likelihood $p(\vx)$, the EnKF propagates $N$ equally weighted particles, denoted as $\z_{t-1}^{1:N}$, from the filtering distribution, $p_{\theta}(\z_{t-1}\vert \x_{1:t-1})$, see Eq.~\eqref{eq:filtering_distribution}, using the transition function to approximate the forecast distribution, $p_{\theta}(\z_t|\x_{1:t-1})$:
\begin{subequations}
    \begin{align}
    p_\theta\left(\z_t \vert \x_{1: t-1}\right) & =\int p_{\theta}(\z_t|\z_{t-1}) p_\theta\left(\z_{t-1} \vert \x_{1: t-1}\right) \mathrm{d} \z_{t-1},\\
    & \approx  \mathcal{N}(\widehat{m}_t, \widehat{C}_t),  \ \cdots \cdots   \  \text{(forecasting step)}
\end{align}
\end{subequations}
where \vspace{-.05in}
\begin{subequations}
    \begin{align}
        & \widehat{m}_t = \frac{1}{N} \sum_{n=1}^N \widehat{\z}_t^n,\\
        & \widehat{C}_t = \frac{1}{N-1} \sum_{n=1}^N\left(\widehat{\z}_t^n-\widehat{m}_t\right)\left(\widehat{\z}_t^n-\widehat{m}_t\right)^{\top},
    \end{align}
\end{subequations}
and $\widehat{\z}_{t}^n=F_\alpha\left(\z_{t-1}^n\right)+\xi_t^n, n \!=\! 1,2, \ldots, N$, denotes the forecast ensemble.

Then, due to the linearity and the Gaussian nature of the emission model, we can recursively obtain the filtering distribution at time step $t$, $p_{\theta}(\z_t \mid \x_{1: t})$:
\begin{subequations}
\label{eq:filtering_distribution}
\begin{align}
&p_{\theta}(\z_t \vert \x_{1: t}) = \mathcal{N}({m}_t, {C}_t),  \ \ \ \cdots \cdots  \ \ \  \text{(filtering step)} \\
&{m}_t = \widehat{m}_t + \widehat{K}_t\left(\x_{t} -H \widehat{m}_t\right), \\
&{C}_t = \widehat{C}_t - \widehat{C}_t  \widehat{K}_t^{\top} \widehat{C}_t ^\top,
\end{align}
\end{subequations}
where $\widehat{K}_t \triangleq \widehat{C}_t H^{\top}(H \widehat{C}_t H^{\top}+R)^{-1}$ represents the Kalman gain. With these steps, the logarithm of the marginal likelihood thus can be approximated as
\begin{subequations} \label{eq: lml}
\begin{align}
    \mathcal{L}_{\mathrm{EnKF}}(\theta) & \triangleq \log p_{\theta}\left(\vx\right)=\sum_{t=1}^T \log p_\theta\left(\x_t \vert \x_{1: t-1}\right) \\
    & = \sum_{t=1}^T \log \int p(\x_{t}|\z_{t}) p_\theta\left(\z_t \vert \x_{1: t-1}\right) \mathrm{d} \z_t\\
    & = \sum_{t=1}^T \log \mathcal{N}\left(H \widehat{m}_t, H \widehat{C}_t H^{\top}+R\right).
\end{align}
\end{subequations}
Leveraging the reparameterization trick \cite{kingma2013auto} in forecast and filtering steps, the autodifferentiable EnKF constructs the map $\theta \!\mapsto\! \mathcal{L}_{\mathrm{EnKF}}(\theta)$ \cite{chen2022autodifferentiable}. Consequently, we can optimize the $\theta$ via gradient descent-based methods. Yet, all existing learning methods, including the autodifferentiable EnKF, exclusively address single-task scenarios and lack suitability for managing multiple tasks. Our focus will shift to exploring continual learning in DSSMs in the next section.

\section{Continual Learning in DSSMs} 
\label{sec: proposed_mothod} 
\begin{figure}[t!]
	\centering
	\footnotesize
	\begin{tikzpicture}[align = center, data/.style={rectangle, draw, fill=gray!20, text width = 1cm}, function/.style={rectangle, draw, fill=cyan!20, text width = 1cm}, transparent/.style={rectangle, text width = 1cm}, node distance=1.5cm]
 	\node[data](x1) {$\vx_{1}$};
        \node[data, right of=x1](x2) {$\vx_{2}$};
        \node[transparent, right of=x2](x3) {$\cdots$};
		\node[data, right of=x3](xJ-1) {$\!\!\vx_{J-1}\!\!$};
		\node[data, right of=xJ-1](xJ) {$\vx_{J}$};
  	\node[function, below of=x1](f1) {$\!\!\textsc{dssm}(\theta)\!\!$};
  	\node[function, below of=x2](f2) {$\!\!\textsc{dssm}(\theta)\!\!$};
  	\node[transparent, right of=f2](f3) {$\cdots$};
		\node[function, below of=xJ-1](fJ-1) {$\!\!\textsc{dssm}(\theta)\!\!$};
		\node[function, below of=xJ](fJ) {$\!\!\textsc{dssm}(\theta)\!\!$};

  	\draw[-latex] (x1) -- (f1);
        \draw[-latex] (f1) -- (x2);
        \draw[-latex] (x2) -- (f2);
        \draw[-latex] (f2) -- (x3);
        \draw[-latex] (f3) -- (xJ-1);
  	\draw[-latex] (xJ-1) -- (fJ-1);
        \draw[-latex] (fJ-1) -- (xJ);
		\draw[-latex] (xJ) -- (fJ);
        
	\end{tikzpicture}
	\caption{Illustration of continual learning in DSSMs}
	\label{fig:task_scenarios}
\end{figure}
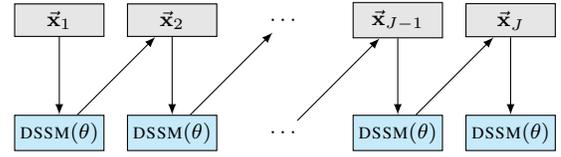

This section elaborates on the main ingredients of our proposed method. As described in Fig.~\ref{fig:task_scenarios}, we consider $J\in \mathbb{N}$ interrelated dynamic modeling tasks, each observed sequentially. The associated datasets cannot be stored due to limited storage memory. Specifically, we denote the observable data of size $T_j$ for the $j$-th dynamic modeling task as $\vx_{j} = \{ \x_{j,t}\}_{t = 1}^{T_j}$, $j = 1,\ldots, J$. Given the observed data, the objective for the DSSM is to continually learn the underlying dynamic systems without experiencing catastrophic forgetting.
This ensures that the learned DSSM can be utilized to make predictions across various dynamic modeling tasks. Additionally, we aim to maintain a constant level of memory consumption for model training, irrespective of the number of tasks.

As mentioned in Section \ref{sec: introduction}, this paper focuses on regularization-based CL methods, of which the general idea involves introducing a weighted regularization term to the original loss to penalize deviations in model parameters when new tasks arise. Various established regularization-based CL methods will be integrated into the DSSM \cite{kirkpatrick2017overcoming,schwarz2018progress,aljundi2018memory,zenke2017continual,li2017learning}, except for variational continual learning (VCL) \cite{nguyen2017variational}. This exclusion is motivated by research findings indicating that learning and inference in the Bayesian neural network in VCL can be computationally expensive \cite{panousis2019nonparametric,theodoridis2020machine}.
In the following subsections, we elaborate on how the regularization-based CL methods empower DSSMs to continually learn the underlying system dynamics. 

\subsection{Learning the First Task} 
We begin with the introduction to the learning of the first task, which involves classic learning and inference in DSSMs, as presented in Section \ref{subsec:enkf}. Specifically, we first evaluate the log-likelihood, see Eq.~\eqref{eq: lml} through the propagation of the emission function using the approximated forecast distribution obtained from EnKF. Then, a recognition network is introduced to infer the initial latent state from input observations \cite{krishnan2017structured,lin2023towards}. More concretely, we approximate the posterior distribution of the initial latent state using the introduced parametric distribution, $q_\phi(\z_{1,0}|\vx_1)$, where $\z_{1,0}$ denotes the initial latent state of the first task. The distribution $q_\phi(\z_{1,0}|\vx_1)$ is assumed to be Gaussian, with mean ($\mu_\phi$) and variance ($\Sigma_\phi$) functions modeled by a recurrent neural network (RNN)-based recognition network with parameters $\phi$. Therefore, the loss function aimed at minimizing for the first task is expressed as follows:
\begin{subequations}
\label{eq:2}
\begin{align}
\!\!\! \mathcal{L} (\vx_1; \theta, \phi) & = \sum_{t=1}^T \log \mathcal{N}\left(H \widehat{m}_t, H \widehat{C}_t H^{\top}+R\right) \label{eq:2_1} \\
&\quad - \operatorname{KL}\left[q_\phi(\z_{1,0}|\vx_1) \| p(\z_{1,0})\right], \label{eq:2_2}
\end{align}
\end{subequations}
where Eq.~\eqref{eq:2_1} represents the model log-likelihood function, optimizing the observation reconstruction capability, and Eq.~\eqref{eq:2_2} corresponds to the Kullback-Leibler (KL) divergence between the posterior distributions and the prior distributions, which serves as a regularizer, ensuring that the posterior distributions do not deviate significantly from the prior distributions, facilitating the learning of the system dynamics \cite{klushyn2021latent}. The prior $p\left(\z_{1,0}\right)$ is assumed to adhere to a known Gaussian distribution. Due to the utilization of this reparameterization trick \cite{chen2022autodifferentiable}, we can optimize the model parameter $\theta$ via a modern optimizer, such as Adam \cite{kingma2014adam}.

\subsection{Continual Learning for Subsequent Tasks}\label{subsec:memory-efficient}

This subsection details various regularization-based continual learning methods for the DSSM to continually learn the model parameters $\theta = (\alpha, \beta)$ in the subsequent tasks  \cite{kirkpatrick2017overcoming,schwarz2018progress,aljundi2018memory,zenke2017continual,li2017learning}. Note that the inference network parameters $\phi$ are task-specific and play a relatively marginal role in DSSM \cite{chen2022autodifferentiable}. Therefore, we do not discuss their continual learning but rather train them independently in each task.

\paragraph{Elastic Weight Consolidation (EWC)}
The fundamental concept of EWC revolves around controlling the deviation of parameters that are deemed crucial for previous tasks \cite{kirkpatrick2017overcoming,schwarz2018progress}. Through the addition of an extra regularization term, the overall loss function for EWC is expressed as:
\begin{align}
\label{eq: EWC}
\!\!\!\! \mathcal{L}  (\vx_{j};\theta, \phi)  \!=\! \mathcal{L}(\vx_{j}; \theta, \phi) \!+\!\sum_{k=1}^{j\!-\!1}\left(\!\frac{\lambda}{2} \sum_i M_{k, i}\!\left(\theta_i\!-\!\theta_{k, i}^*\right)^2\!\right),
\end{align}
where $\lambda$ is a weighting hyperparameter that represents the importance of previous tasks, and $M_{k}$ is the diagonal Fisher information matrix for the $k$-th task evaluated at $\theta^*_{k}$, which quantifies the importance of each model parameter with respect to the corresponding task \cite{rissanen1996fisher}. The parameter learned from the $k$-th task is denoted as $\theta_{k}^*$, and the subscript $i$ denotes the $i$-th parameter in the model. 

However, the number of quadratic regularization terms grows with the number of tasks in the vanilla EWC \cite{kirkpatrick2017overcoming}, resulting in escalating memory and computation requirements.  Online-EWC addresses this limitation by defining the following objective function:
\begin{align}
\label{eq: OEWC}
\!\!\!\! \mathcal{L}  (\vx_{j} ;\theta, \phi) \!=\! \mathcal{L}(\vx_{j}; \theta, \phi) \!+\!\frac{\lambda}{2} \sum_i \left(\tilde{M}_{j\!-\!1, i}\left(\theta_i\!-\!\theta_{j\!-\!1, i}^*\right)^2\right).
\end{align}
Unlike the vanilla EWC, see Eq.~\eqref{eq: EWC}, the quadratic regularization term in Online-EWC, see Eq.~\eqref{eq: OEWC}, is a single moving sum on each related parameter, i.e., $\tilde{M}_{j} \!=\! \gamma\tilde{M}_{j-1} + M_{j}$, where  $\tilde{M}_{1} = M_{1}$ and  $\gamma$ ($\gamma \!\leq\! 1$) is a hyperparameter governing the contribution of each previous task. Notably, after training the $(j\!-\!1)$-th task ($j>1$), the network parameters are already optimal for all preceding tasks. Consequently, there should be only one regularization term anchored at the parameters learned from the latest task (which is denoted as $\theta_{j-1}^*$ in Eq.~\eqref{eq: OEWC}). 
Due to the advantageous on-memory computational characteristics of Online-EWC, the subsequent sections of this paper solely focus on Online-EWC, which will be referred to as EWC throughout the remainder of this paper for simplicity.

\paragraph{Memory Aware Synapses (MAS)}
Similar to EWC, the regularization term in MAS is derived from the gradient of the learned function, and the importance of model parameters can be updated in an online fashion \cite{aljundi2018memory}. Specifically, the loss function of MAS is expressed as:
\begin{align}
\label{eq: MAS}
\!\!\!\!\mathcal{L}  (\vx_{j} ;\theta, \phi) \!=\!  \mathcal{L} (\vx_{j}; \theta, \phi)  \!+\!\lambda \sum_i \left(\Omega_{j-1, i}\left(\theta_i \!-\! \theta_{j\!-\!1, i}^*\right)^2\right),
\end{align}
where $\lambda$ is the hyperparameter with a similar role as in EWC; $\Omega_{j} = \frac{1}{T_j} \sum_{t=1}^{T_j} \parallel g(\z_{j, t}) \parallel$ represents the parameter importance matrix and controls the deviation of model parameters, where $g(\z_{j, t})$ is the gradient of the transition evaluated at each state $\z_{j,t}$ with respect to the parameter $\theta$. 

\paragraph{Synaptic Intelligence (SI)}
In comparison to previous methods, the update process for the importance of SI is more complicated \cite{zenke2017continual}. The loss function of SI is given by:
\begin{align}
\label{eq: SI}
\!\!\!\!\mathcal{L}  (\vx_{j} ;\theta, \phi) \!=\!  \mathcal{L} (\vx_{j}; \theta, \phi) \!+\! \lambda \sum_i \left(\Lambda_{j-1, i}\left(\theta_i\!-\!\theta_{j-1, i}^*\right)^2\right),
\end{align}
where $\lambda$ ($\lambda \!\leq\! 1$) is a hyperparameter that balances the influence between tasks,  and $\Lambda_{j-1}$ denotes the importance matrix for the previous $j\!-\!1$ tasks and can be represented as:
\begin{equation}
\label{eq: lambda}
    \Lambda_{j-1, i} = \sum_{k=1}^{j-1} \frac{\omega_{k,i}}{(\Delta_{k,i})^2 + \epsilon},
\end{equation}
where $\Delta_{k} \!=\! \theta_{k}[I] - \theta_{k}[0]$ represents the deviation of parameters after total $I$ iterations of training on the $k$-th task, and $\omega_{k}$ indicates the importance of each parameter to the loss and is obtained by the product of two gradients, i.e.,  $\omega_{k} = \frac{\partial \mathcal{L}(\vx_{k})}{\partial \theta_{k}} \frac{\partial \theta_{k}}{\partial t}$. Note that here $\omega_{k}$ is updated in every iteration, while $\Delta_{k}$ is only updated after $I$ iterations. The parameter $\epsilon$ serves as a damping parameter in the case where $\Delta_{k} \rightarrow 0$, and it is set to $\epsilon = 0.01$ by default \cite{zenke2017continual}.

\paragraph{Learning without Forgetting (LwF)}
The regularization utilized in LwF takes a different approach from the aforementioned methods, which is often termed functional regularization rather than parameter regularization \cite{li2017learning}. 
Unlike the original LwF, which takes a distillation loss as the regularization term for classification problems, we use the mean squared error (MSE) loss instead given the observations are continuous. Specifically, the loss function with LwF is given by: 
\begin{align}
\label{eq: LwF}
& \mathcal{L}  (\vx_{j} ;\theta, \phi) \!=\! \mathcal{L} (\vx_{j}; \theta, \phi) \!+\! \frac{\lambda}{j\!-\!1} \sum_{k=1}^{j\!-\!1} \mathcal{L}_{\textsc{mse}}\left(\hat{\x}_{k}(\theta), \hat{\x}_{k}(\theta_{j-1}^*)\right),
\end{align}
where $\hat{\x}_{k}(\theta_{j-1}^*)$ denotes the forecast observation sequence for the $k$-th task, $k = 1, 2, \dots j-1$, using the DSSM with parameter $\theta_{j-1}^*$, while  $\hat{\x}_{k}(\theta)$ is the forecast observation generated with the updated model parameters, and the $\mathcal{L}_{\textsc{mse}}\left(\cdot, \cdot \right)$ represents the standard MSE loss \cite{theodoridis2020machine}. 
The length of the forecast observation sequence is the same as the test dataset of each task.

\subsection{Computational and Memory Cost Analysis}
\label{subsec: computation}

The overall computational and memory cost of the CLDSSM arises from the autodifferentiable EnKF and the additional regularization. The cost associated with the autodifferentiable EnKF is task-independent, with a memory cost of $\mathcal{O}(d_{z}N)$ and a computational cost of $\mathcal{O}({d_{z}}{d_{x}}N)$ for the Kalman Gain evaluation \cite{roth2017ensemble}. The complexity of ensemble forward propagation depends on the structure of the deep neural network-based transition function within the DSSM.

Regarding the regularization part, the memory costs for EWC, MAS, and SI are consistent with $\mathcal{O}({d_{z}}^{2})$, while LwF incurs a memory cost of $\mathcal{O}({d_{z}}T)$, where $T$ denotes the size of the forecast data. In terms of computational costs, both EWC and MAS scale as $\mathcal{O}({d_{z}}^{2})$, while SI scales as $\mathcal{O}({d_{z}}^{3})$, and the cost of LwF is dominated by the computation of MSE part, which scales as $\mathcal{O}({d_{z}}T)$. 

To be more specific, under general conditions where $d_{z} \ll T$, the memory costs of different methods can be ranked as EWC=MAS=SI$<$LwF, while the computational costs are EWC=MAS$<$SI$<$LwF. The parameter regularization methods, including EWC, MAS, and SI, control the deviation of parameters through different gradients and are expected to be effective in retaining data features under multi-task training. Meanwhile, the functional regularization method, LwF, constrains parameter training via the MSE function, likely resulting in better performance in numerical evaluations during experiments. Moreover, although the LwF method incurs significantly higher storage usage, its structure and computation are more concise and succinct compared to the other methods\cite{li2017learning}.

It is also noteworthy that the computational and memory costs of all regularization methods are independent of the number of tasks and the size of training data. Consequently, the corresponding CLDSSM is highly efficient and does not exhibit the issue of computational and memory escalation as in conventional DSSMs. 

\section{Experiments} \label{sec: experiment}
We assess the effectiveness of the proposed CLDSSMs across two real-world datasets, demonstrating their ability to overcome catastrophic forgetting issues while keeping a constant memory and computational cost.  The main evaluation metric employed for this assessment is the MSE of the model observation predictions, i.e.,
\begin{equation}
\text{MSE} \triangleq \mathbb{E}\left[\| \x-\widehat{\x}\|_2^2\right],
\end{equation}
where $\x$ represents the ground truth, and $\widehat{\x}$ is the predictive output. For each individual dataset, the training hyperparameter of each method is consistent across all tasks, indicating that each task is assigned equal importance. In our experiment, we set $\lambda_{\text{EWC}} = 1000$ for EWC, $\lambda_{\text{MAS}} = 800$ for MAS, $\lambda_{\text{SI}} = 1$ for SI and $\lambda_{\text{LwF}} = 1$ for LwF. The optimizer used is Adam \cite{kingma2014adam}, with a learning rate of 0.005.

\subsection{Power Consumption Dataset} \label{subsec:power_datasets}
We first evaluate CLDSSMs using a real-world power consumption dataset obtained from Tetouan city \cite{salam2018comparison}. The dataset spans the year 2017 with a 10-minute time interval and comprises 52,416 samples with five-dimensional control inputs and three-dimensional observations. The observations include recorded power consumption in three distribution networks in Tetouan, influenced by diverse inputs such as temperature, humidity, wind speed, diffuse flow, and general diffuse flow. 

To assess the proficiency of CLDSSMs in continually learning multiple tasks, we divide the dataset into four segments chronologically, corresponding to four learning sequential tasks. Subsequently, we randomly select 1800 successive samples from each segment to form 32 training target sequences with a sequence length of $T = 50$. The remaining 200 samples in each segment are reserved as the test data for predictive analysis.

We compare the various CLDSSMs with the baseline DSSM, and Table~\ref{tab1} reports the prediction MSE, where the averaged MSE is evaluated using all the test data from the trained tasks. Thus, the averaged MSE can indicate the effectiveness of our approach in overcoming catastrophic forgetting. Our results highlight the substantial performance improvement of CLDSSMs over DSSM, especially when dealing with multiple tasks. Notably, the MSE values sharply increase during the 3rd and 4th tasks when training with the baseline DSSM, indicating significant challenges in transferring parameters from old to new tasks. In contrast, all of our proposed CLDSSMs consistently maintain lower MSE values, showcasing their ability to mitigate catastrophic forgetting. Furthermore, the smaller standard deviations in CLDSSMs results indicate enhanced stability compared to the baseline. 

\begin{figure}[t]
    \centering
    \begin{minipage}{0.45\linewidth}
        \centering
        \includegraphics[width=1\linewidth]{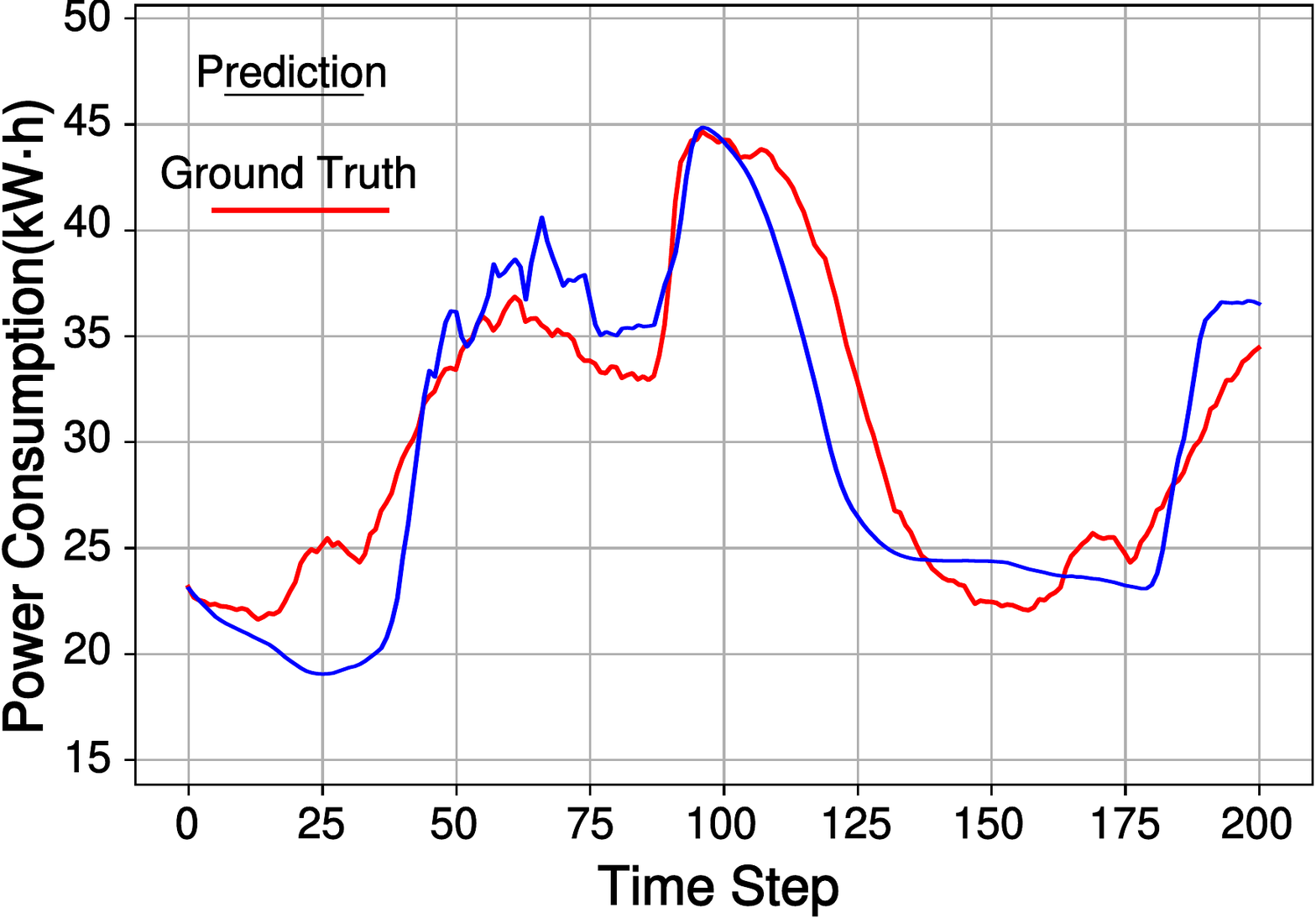}
        \subcaption{DSSM-first task only}
        \label{fig1_0}
    \end{minipage}
    \begin{minipage}{0.45\linewidth}
        \centering
        \includegraphics[width=1\linewidth]{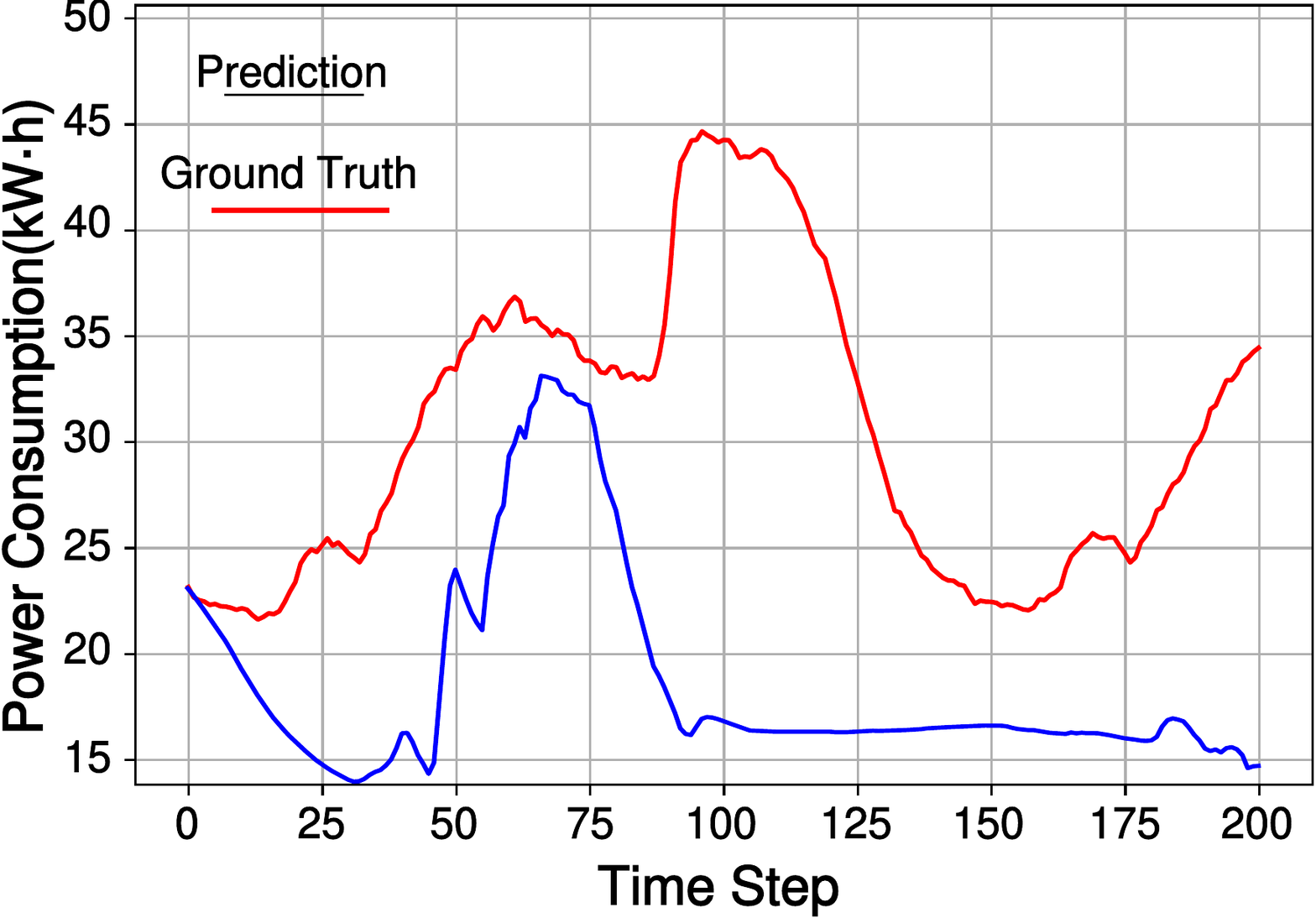}
        \subcaption{DSSM}
        \label{fig1_1}
    \end{minipage}

    \begin{minipage}{0.45\linewidth}
        \centering
        \includegraphics[width=1\linewidth]{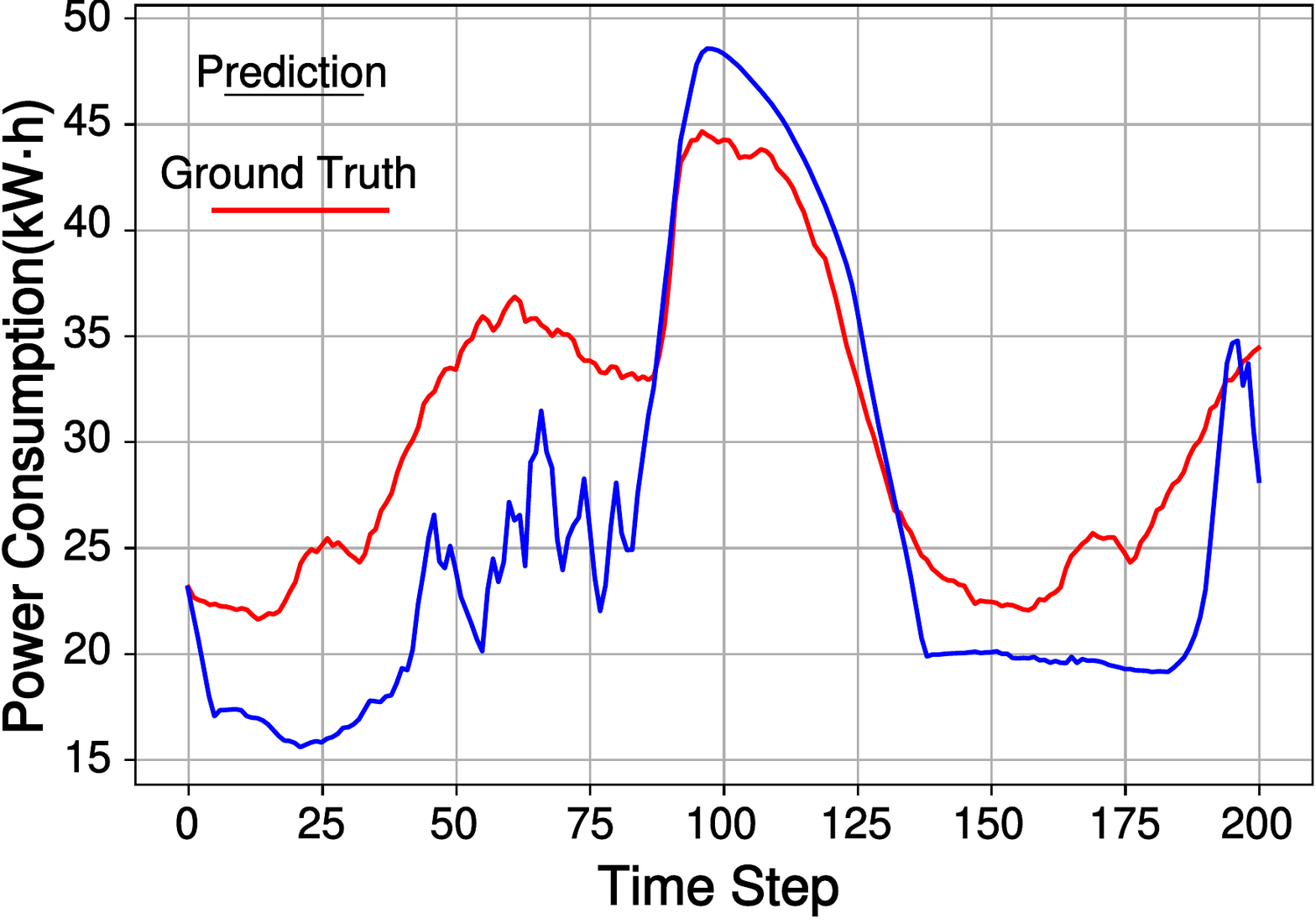}
        \subcaption{CLDSSM-EWC}
        \label{fig1_2}
    \end{minipage}
    \begin{minipage}{0.45\linewidth}
        \centering
        \includegraphics[width=1\linewidth]{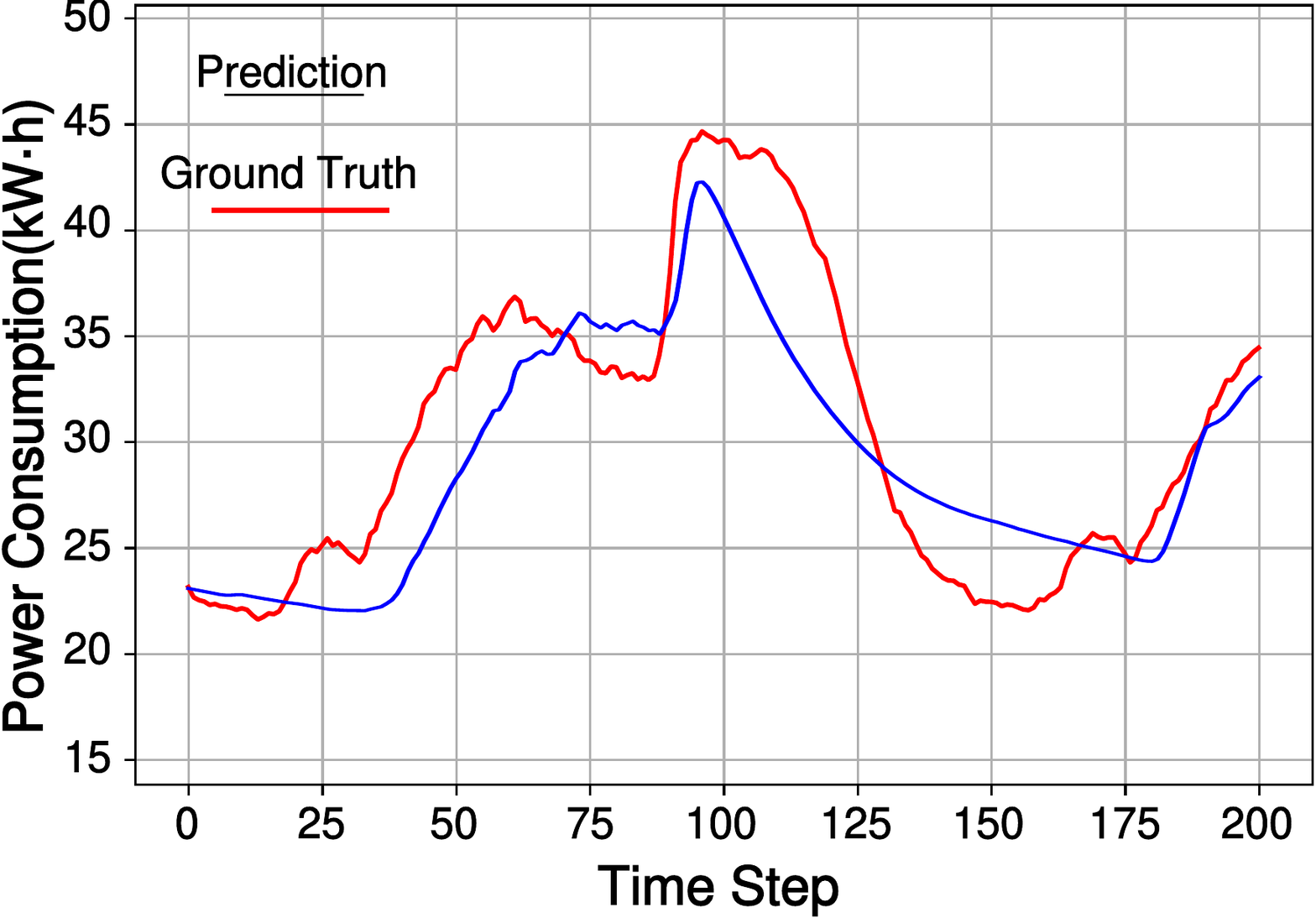}
        \subcaption{CLDSSM-MAS}
        \label{fig1_3}
    \end{minipage}

    \begin{minipage}{0.45\linewidth}
        \centering
        \includegraphics[width=1\linewidth]{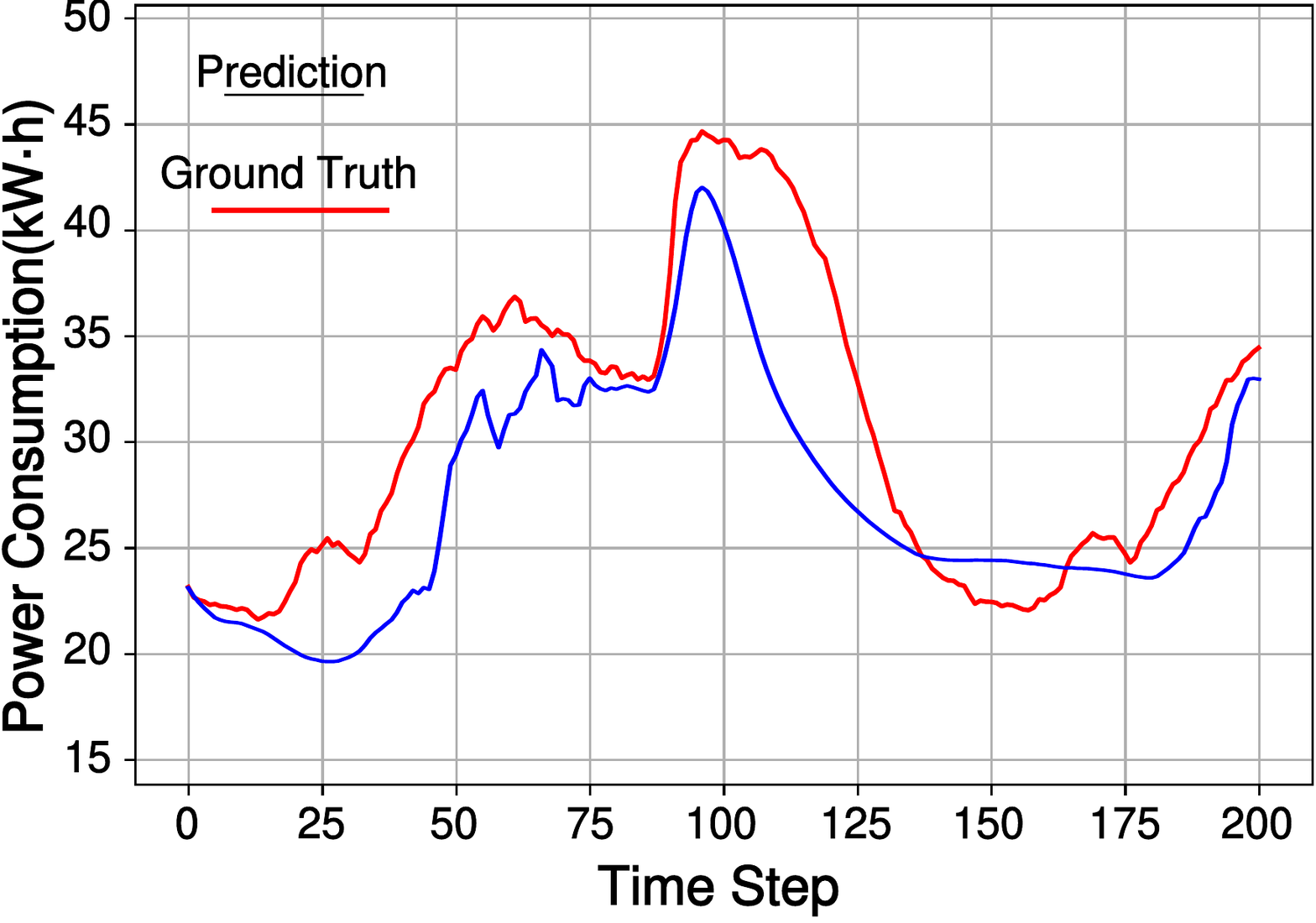}
        \subcaption{CLDSSM-LwF}
        \label{fig1_4}
    \end{minipage}
    \begin{minipage}{0.45\linewidth}
        \centering
        \includegraphics[width=1\linewidth]{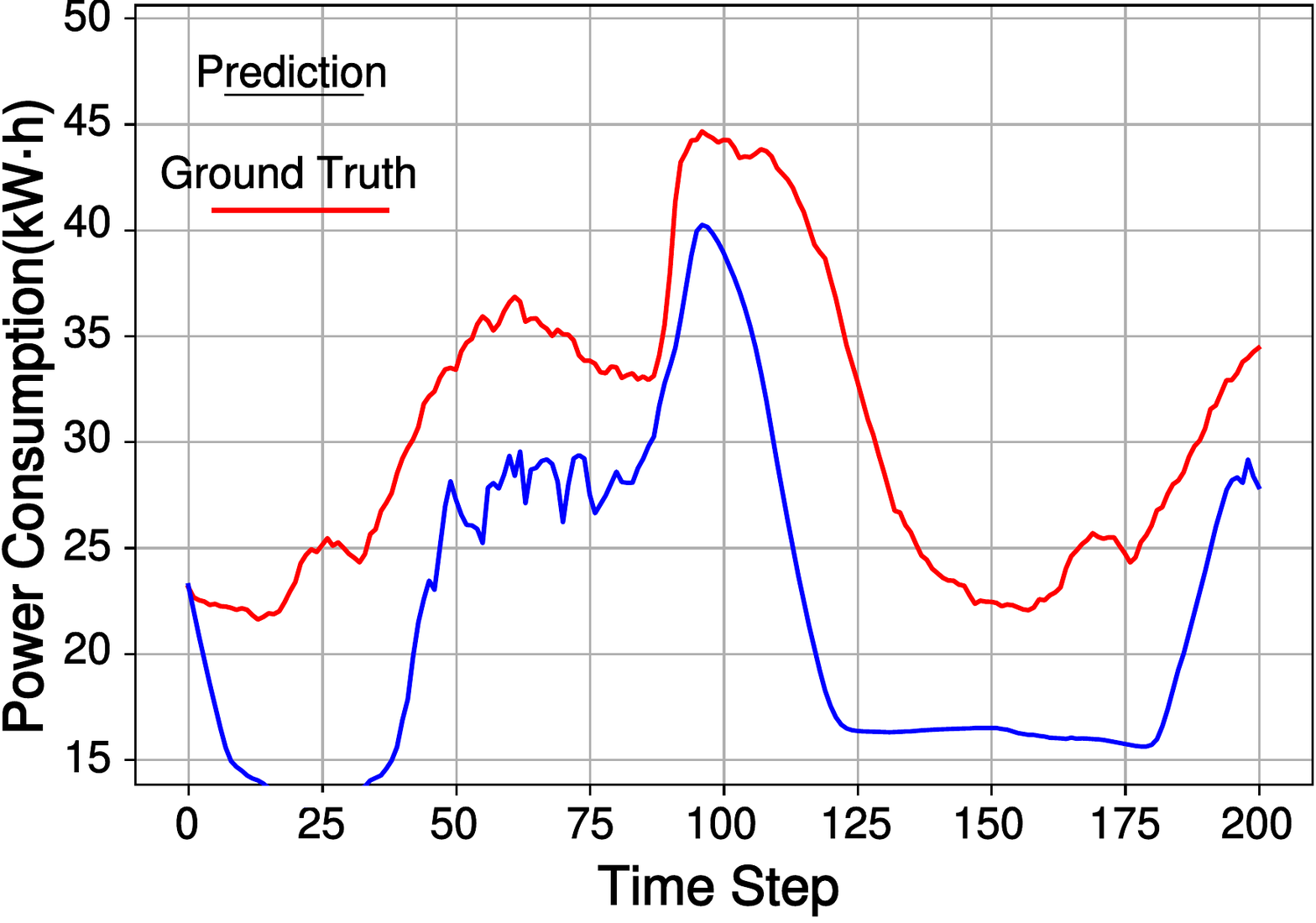}
        \subcaption{CLDSSM-SI}
        \label{fig1_5}
    \end{minipage}
    \caption{Prediction results of DSSM and CLDSSMs in the 1st task of the \textsc{power consumption} dataset. 
    }
    \label{fig1}
\end{figure}

The more detailed results of various models are depicted in Fig.~\ref{fig1}, where the predictions are made for a total of 200 time steps immediately following the training sequence. Fig.~\ref{fig1_0} displays the result of the baseline DSSM when trained solely on the $1$st task. Subsequent figures (Figs.~\ref{fig1_0}--\ref{fig1_5}) depict predictions for the 1st task after training all 4 tasks using the respective labeled baseline and CLDSSMs.  Fig.~\ref{fig1_1} shows that the observation predictions from the baseline DSSM exhibit a notable disparity from the ground truth, demonstrating its incapability in predicting the trajectory beyond 100 time steps. In contrast, the prediction results of all CLDSSMs consistently align with the ground truth across all 200 time steps, even with a slight mismatch. Notably, the prediction of LwF exhibits superior overall performance with the lowest MSE value after training all tasks, while EWC, MAS, and SI also closely follow the observed rise-and-fall trajectory patterns.  The performance difference can be attributed to the different rationales in continual learning methods, as discussed in Section \ref{subsec: computation}. For example, the EWC method controls parameter deviations, making it adept at capturing trajectory features, while the LwF method utilizes the MSE function to achieve a better holistic result.

\begin{table}[t]
\caption{Averaged prediction MSE of the \textsc{power consumption} dataset. The mean and standard deviation of the prediction results are shown. \vspace{-.1in}}
\begin{center}
\setlength{\tabcolsep}{1.6mm}{
\begin{tabular}{ccccc}
\toprule
\textbf{}&{\textbf{1st Task}}&{\textbf{2nd Task}}&{\textbf{3rd Task}}&{\textbf{4th Task}} \\
\midrule
\textbf{DSSM} & 26.15$\pm$1.12 & 56.21$\pm$5.35 & 299.32$\pm$19.74 & 463.36$\pm$26.49 \\
\midrule 
\textbf{EWC} & \textbf{26.03}$\pm$0.98 & \textbf{25.65}$\pm$3.76 & 58.41$\pm$4.84 & 77.01$\pm$7.52 \\
\textbf{MAS} & 27.28$\pm$1.06 & 29.26$\pm$4.04 & 55.64$\pm$5.18 & 71.87$\pm$6.98 \\
\textbf{LwF} & 26.39$\pm$1.10 & 27.19$\pm$3.27 & \textbf{51.77}$\pm$4.63 & \textbf{69.65}$\pm$7.26 \\
\textbf{SI} & 27.94$\pm$1.21 & 36.44$\pm$4.39 & 64.70$\pm$6.59 & 101.14$\pm$9.63 \\
\bottomrule
\end{tabular}} 
\label{tab1}
\end{center}
\end{table}

\subsection{Weather Dataset}
In this subsection, we conduct another evaluation using a real-world weather dataset comprising 5844 samples with four-dimensional control inputs and one-dimensional observations. This dataset captures the weather information in London from 2005 to 2020, encompassing input features such as cloud cover, sunshine conditions, global radiation, and precipitation. The observed variable is represented by temperature measured in degrees Celsius. Similar to the approach outlined in Section~\ref{subsec:power_datasets}, we partition the dataset into four segments, each containing 1461 samples spanning four years and serving as an individual task. 
The sequence length is defined as $T = 50$, and the batch size is set to 22, preserving data for approximately one year as the ground truth for predictions.

\begin{figure}[t!]
    \centering
    \begin{minipage}{0.45\linewidth}
        \centering
        \includegraphics[width=1\linewidth]{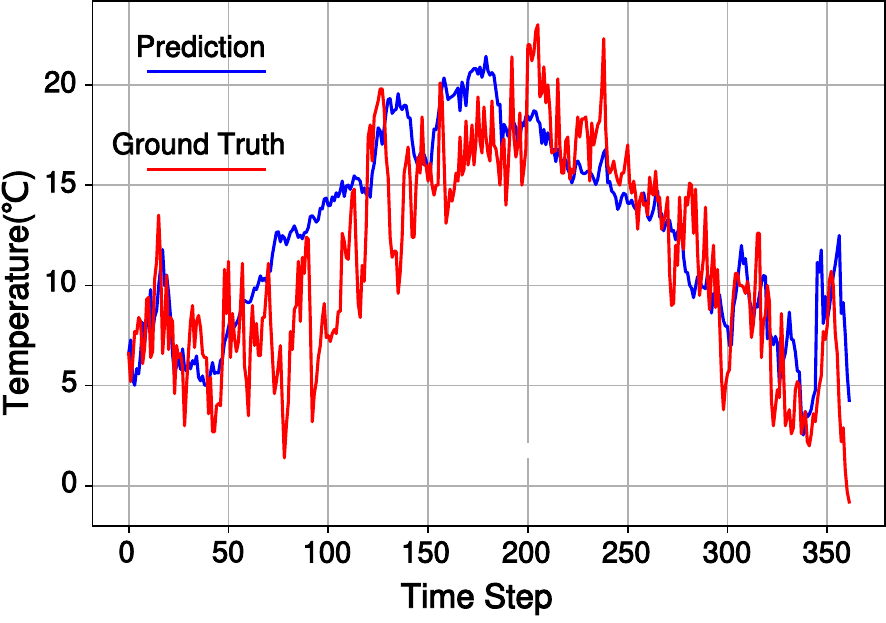}
        \subcaption{DSSM-first task only}
        \label{fig2_0}
    \end{minipage}
    \begin{minipage}{0.45\linewidth}
        \centering
        \includegraphics[width=1\linewidth]{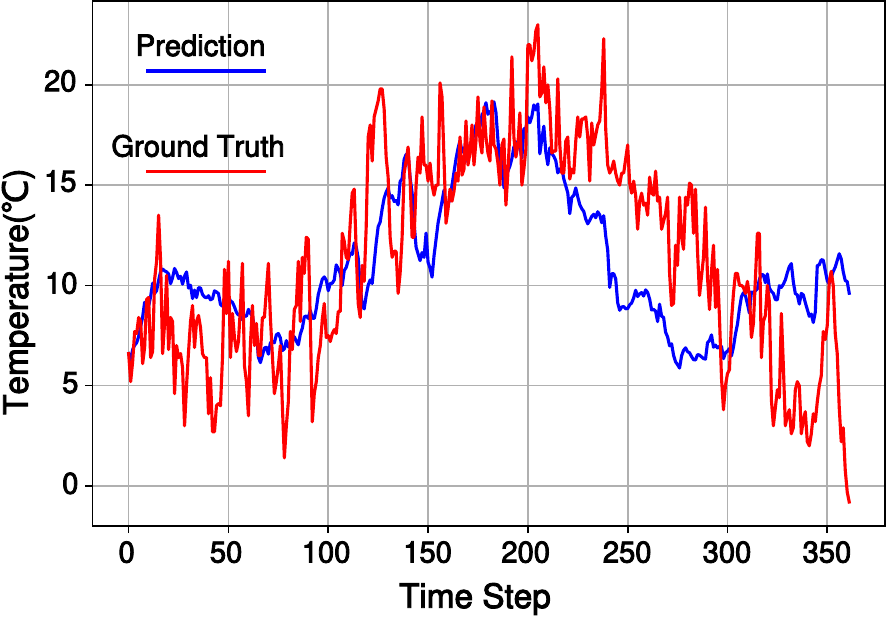}
        \subcaption{DSSM}
        \label{fig2_1}
    \end{minipage}

    \begin{minipage}{0.45\linewidth}
        \centering
        \includegraphics[width=1\linewidth]{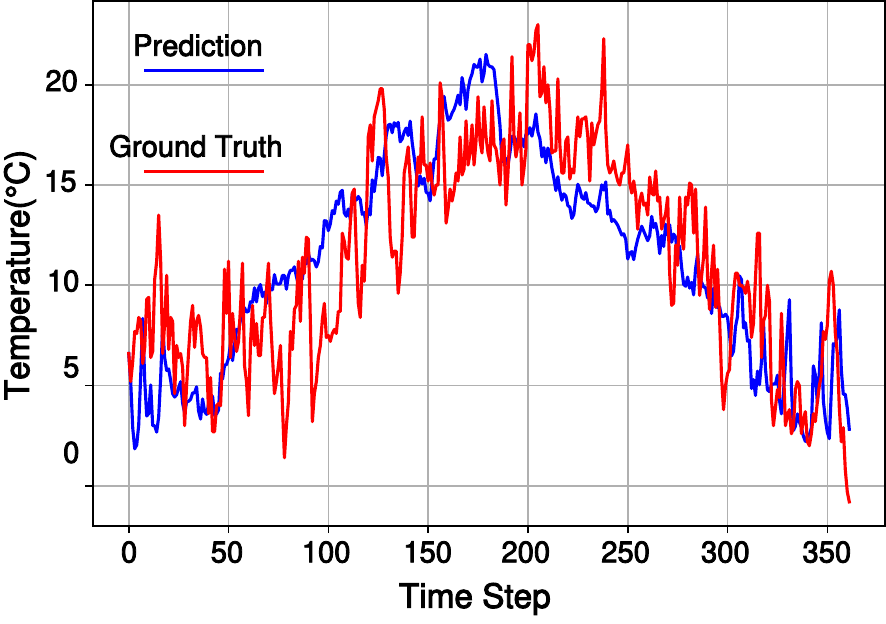}
        \subcaption{CLDSSM-EWC}
        \label{fig2_2}
    \end{minipage}
        \begin{minipage}{0.45\linewidth}
        \centering
        \includegraphics[width=1\linewidth]{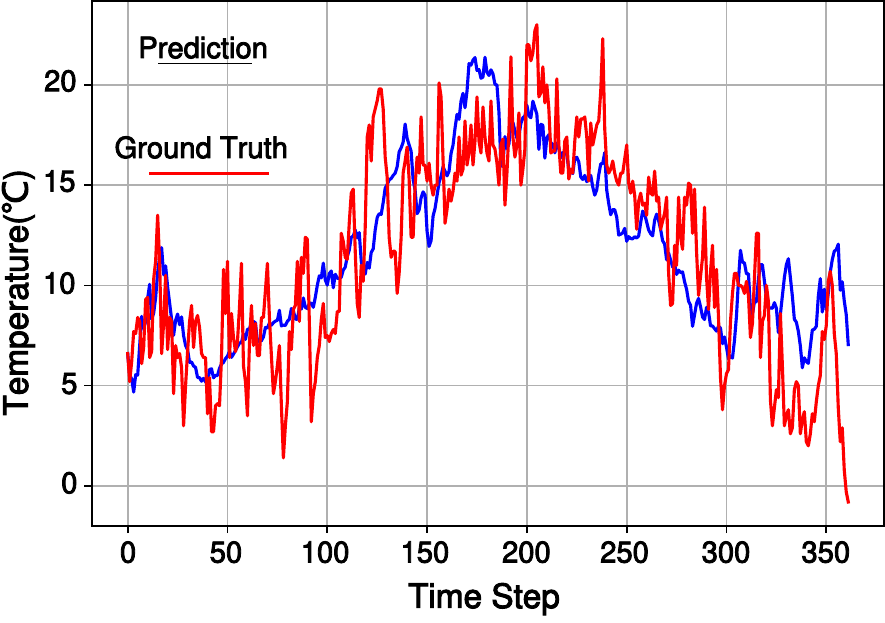}
        \subcaption{CLDSSM-LwF}
        \label{fig2_3}
    \end{minipage}
    \caption{Prediction results of DSSM and CLDSSMs in the 1st task of the \textsc{weather} dataset. 
    }
    \label{fig2}
\end{figure}

Table~\ref{tab2} presents an overview of different model prediction results for the weather data, evaluated by the MSE between ground truth and predictions across all tasks. Our proposed CLDSSMs exhibit notable effects in mitigating catastrophic forgetting, consistently exceeding the baseline with lower MSE values and more stable standard deviations. Similar to the previous subsection, we also depict in Fig.~\ref{fig2} the prediction results of all the models in the 1st task after four tasks of learning. The 361 test data points come immediately after the training data, with a predicted period spanning approximately one year. Notably, the results of CLDSSMs outperform the baseline, with a significant improvement in prediction accuracy at later steps. This is evident in Fig.~\ref{fig2_1}, where the predictions of the DSSM struggle to converge with the ground truth, eventually failing to follow the trajectory after 200 time steps. In general, similar to the results in the last subsection, the LwF-based CLDSSM demonstrates the lowest prediction MSE after all the training, indicating its ability to achieve the best prediction results.

\begin{figure}[t!]
    \centering
    \begin{minipage}{0.45\linewidth}
        \centering
        \includegraphics[width=1\linewidth]{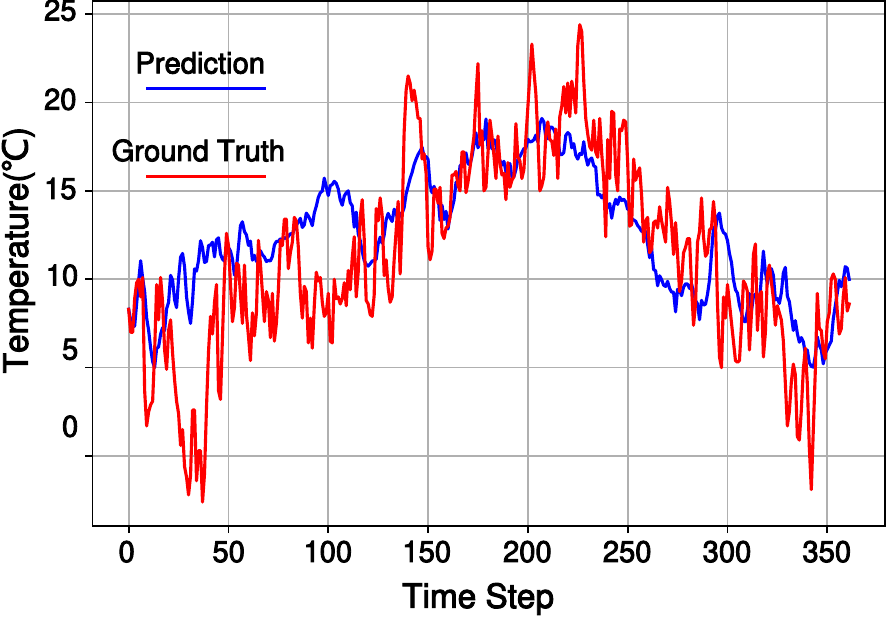}
        \subcaption{DSSM}
        \label{fig3_0}
    \end{minipage}
    \begin{minipage}{0.45\linewidth}
        \centering
        \includegraphics[width=1\linewidth]{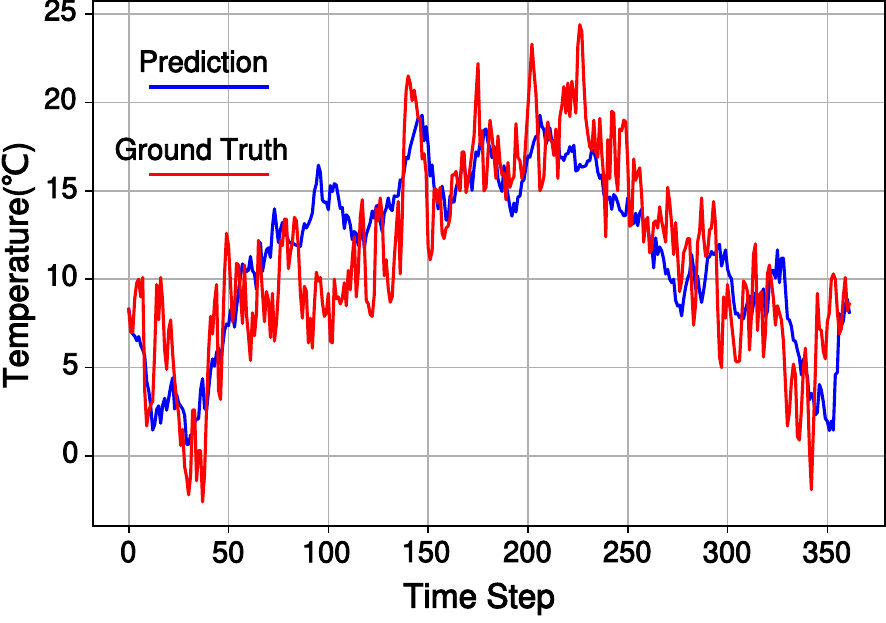}
        \subcaption{CLDSSM-EWC}
        \label{fig3_1}
    \end{minipage}
    \caption{Prediction results of DSSM and CLDSSM-EWC in the 4th task of the \textsc{weather} dataset. 
    }
    \label{fig3}
\end{figure}

Lastly, it is noteworthy that our experiments and findings indicate that CLDSSMs demonstrate accelerated convergence when confronted with new tasks during training, surpassing even the predictive performance of baseline DSSM on the current task. This phenomenon is exemplified in Fig.~\ref{fig3}, where Fig.~\ref{fig3_0} and Fig.~\ref{fig3_1} depict prediction results for the baseline and EWC respectively. Both results represent predictions for the 4th task obtained after training across all 4 tasks. It is evident that the trajectory of CLDSSM-EWC performs globally better and with greater accuracy. Moreover, our empirical results show that the training MSE value for CLDSSM-EWC drops to 18 after 400 epochs of training, while baseline DSSM only reaches a value of around 23 under the same conditions. This notable improvement can be attributed to efficiently handling pre-existing knowledge from previous tasks, facilitating faster adaptation to new ones. In contrast, the absence of regularization in baseline DSSM results in slower training on new tasks and insufficient learning within the same epoch, leading to unsatisfactory test results on all tasks. The ability of such task knowledge extraction in the CLDSSMs highlights the learning efficiency of CLDSSMs from another perspective. 

\begin{table}[t!]
\caption{Averaged prediction MSE of the \textsc{weather} dataset. The mean and standard deviation of the prediction results are shown. \vspace{-.1in}}
\begin{center}
\setlength{\tabcolsep}{1.6mm}{
\begin{tabular}{ccccc}
\toprule
\textbf{}&{\textbf{1st Task}}&{\textbf{2nd Task}}&{\textbf{3rd Task}}&{\textbf{4th Task}} \\
\midrule 
\textbf{DSSM} & 11.31$\pm$0.67 & 13.12$\pm$1.19 & 16.21$\pm$1.83 & 18.16$\pm$1.89 \\
\midrule  
\textbf{EWC} & 12.01$\pm$0.71 & \textbf{11.73}$\pm$0.95 & 13.79$\pm$1.64 & 15.34$\pm$1.47 \\
\textbf{MAS} & \textbf{10.98}$\pm$0.59 & 11.95$\pm$1.03 & \textbf{13.56}$\pm$1.36 & 15.67$\pm$1.72 \\
\textbf{LwF} & 11.74$\pm$0.69 & 12.29$\pm$0.97 & 13.85$\pm$1.29 & \textbf{14.79}$\pm$1.41 \\
\textbf{SI} & 11.48$\pm$0.54 & 12.94$\pm$1.33 & 14.62$\pm$1.35 & 16.73$\pm$1.64 \\
\bottomrule
\end{tabular}}  
\label{tab2}
\end{center}
\end{table}

\section{Conclusion} \label{sec: conclusion}
This paper introduces a novel class of models called continual learning deep state-space models (CLDSSMs), specifically designed to address the challenge of continual learning in DSSMs. The proposed CLDSSMs demonstrate computational and memory efficiency by incorporating regularization-based methods and leveraging training data from the most recent task. Experiments conducted on various real-world datasets showcase the effectiveness of CLDSSMs, highlighting their proficiency in multi-task forecasting and superiority in training new tasks. Furthermore, CLDSSMs with EWC and MAS methods exhibit the lowest costs in terms of memory storage and computational complexity, while LwF produces the best forecasting results, achieving the lowest MSE after training on all tasks. These results collectively underscore the excellent performance of CLDSSMs in applications of multiple dynamic systems modeling.


\bibliographystyle{IEEEtran}
\bibliography{ref-cldssm.bib}

\end{document}